%% file: main.tex
\documentclass[runningheads]{llncs}
\usepackage{graphicx}

\usepackage{subcaption}

\usepackage{times}
\usepackage{soul}
\usepackage{url}
\usepackage[hidelinks]{hyperref}
\usepackage[utf8]{inputenc}
\usepackage{amsmath}
\usepackage{booktabs}
\usepackage{algorithm}
\usepackage{algorithmic}
\begin{document}
\title{Eterna is Solved}

\author{Tristan Cazenave}

\institute{LAMSADE, Université Paris Dauphine - PSL, CNRS, Paris, France}

\maketitle              % typeset the header of the contribution

\input{content}

\newpage

\bibliographystyle{splncs04}
\bibliography{main}

\end{document}

%% file: content.tex
\begin{abstract}
RNA design consists of discovering a nucleotide sequence that folds into a target secondary structure. It is useful for synthetic biology, medicine, and nanotechnology. We propose Montparnasse, a Multi Objective Generalized Nested Rollout Policy Adaptation with Limited Repetition (MOGNRPALR) RNA design algorithm. It solves the Eterna benchmark.
\end{abstract}

\section{Introduction}

The design of molecules with specific properties is an important topic for research related to health. The RNA design problem, also named the Inverse RNA Folding problem, is a difficult combinatorial problem. This problem is important for scientific fields such as bioengineering, pharmaceutical research, biochemistry, synthetic biology, and RNA nanostructures
\cite{portela2018unexpectedly}.

RNA is involved in many biological functions. Synthetic RNA can be easily produced \cite{Reese2005} and has many applications in synthetic biology, as well as in drug design with the building of riboswitches and ribozymes.

RNA design consists of finding a nucleotide sequence that folds into a desired target structure. Eterna is a standard benchmark for RNA design algorithms. Many algorithms have been applied to this problem over the years. However, none have successfully solved all the Eterna problems. This paper presents a simple algorithm that solves the Eterna benchmark.

RNA molecules are long molecules composed of four possible nucleotides. Molecules can be represented as strings composed of the four characters A (Adenine), C (Cytosine), G (Guanine), and U (Uracil). For RNA molecules of length N, the size of the state space of possible strings is exponential in N. It can be very large for long molecules. The sequence of nucleotides folds back on itself to form what is called its secondary structure. It is possible to find in polynomial time the folded structure of a given sequence. However, the opposite, which is the Inverse RNA Folding problem, is hard \cite{bonnet2020designing}. 

RNA functions are determined by its tertiary structure. The secondary structure is used to determine the tertiary structure according to the base pairing interactions. The bonds between two nucleotides are given by the six possible base pairs (CG, GC, AU, UA, UG, GU). The dot-bracket notation is used to represent the secondary structure, the opening and closing brackets represent the base pairs, and the dots represent the unbounded sites.

The paper is organized as follows: the second section is about previous attempts at designing RNA. The third section presents the algorithms used in Montparnasse. The fourth section details the experimental results.

\section{Previous RNA Design Work}

We start with an overview of previous attempts at RNA design. We then describe in more detail the GREED-RNA program and previous Monte Carlo Search approaches.

\subsection{Previous Attempts at Solving Inverse RNA Folding}

The algorithms used in these methods include adaptive
random walk in RNAinverse \cite{hofacker1994fast}, stochastic local search for RNA-SSD \cite{andronescu2004new} and INFO-RNA \cite{busch2006info}, genetic and evolutionary
algorithms such as MODENA \cite{taneda2015multi} and
aRNAque \cite{merleau2022arnaque}.

 The recent RNA design book \cite{churkin2024rna} contains many papers on RNA design for both the secondary structure and the tertiary structure.

\subsection{GREED-RNA}

We focus here on GREED-RNA \cite{lozano2024simple} as it is a very recent and state-of-the-art program for Eterna. We will compare to GREED-RNA in the experimental results section.

GREED-RNA uses greedy initialization: all base pairs are initialized as GC or CG, and
unpaired positions are all initialized as A. In the early stage of search, it also uses greedy mutation that randomly chooses between GC and CG at random positions. Later, it uses random mutation that randomly replaces base pairs with any other base pair and unpaired nucleotides with any nucleotide.

Sequences are sorted using Multi Objective evaluations. The first objective is their value in base pair distance (BPD), then the Hamming Distance, the probability over ensemble, the partition function, the ensemble defect, and the GC-content distance.

Restarts are performed when the stagnation counter reaches a threshold. It also takes sequences from a pool of sorted sequences.

\subsection{Monte Carlo Search}

Early attempts used UCT combined with local search to address RNA design \cite{yang2017rna}. The program was not tested on the Eterna benchmark.

The NEMO program used Nested Monte Carlo Search \cite{CazenaveIJCAI09} with a handmade playout policy to generate RNA that were further optimized with local search. It could solve 95 of the 100 problems of Eterna.

MCTS and learning from self-play were also used for RNA design \cite{obonyo2022designing,obonyo2024rna,obonyo2024self}.

Generalized Nested Rollout Policy Adaptation (GNRPA) \cite{Cazenave2020GNRPA} was applied to Inverse RNA Folding \cite{Cazenave2020Inverse} also to solve 95 problems. It was further refined with the learning of a prior for the policy using either transformers \cite{cazenave2024monte} or statistics on solved problems \cite{cazenave2024learning}.

GNRPA with Limited Repetitions (GNRPALR) \cite{cazenave2024generalized} was also applied to Inverse RNA Folding with success. The principle is to avoid too deterministic policies by stopping iterations when the same sequence is found a second time at a given level of GNRPA.

\section{Montparnasse}

In this section, we describe the algorithms we have developed for RNA design in the Montparnasse framework. We start with Multi Objective Greedy Randomized Local Search (MOGRLS) which is a simplification of GREED-RNA. We then explain a modification of this algorithm we call Progressive Narrowing (PN) that makes use of restarts and selects the most promising sequences among a set of partially optimized sequences. The Progressive Narrowing algorithm is tuned using a search for the best parameters. We end this section with Multi Objective Generalized Nested Rollout Policy Adaptation with Limited Repetitions (MOGNRPALR), the MCTS algorithm we propose for RNA design.

\subsection{Multi Objective Greedy Randomized Local Search}

MOGRLS is a simplification of GREED-RNA that gives better results on difficult problems. It is a simple algorithm described in Algorithm \ref{AlgoMOGRLS}.

\begin{algorithm}
\caption{MOGRLS}
\begin{algorithmic}[1]
\label{AlgoMOGRLS}
\STATE{MOGRLS ($targetStructure$)}
\begin{ALC@g}
\STATE{$bestSequence \leftarrow GenerateInitialSequence (targetStructure)$}
\WHILE{True}
\IF{$nevals < 500$}
\STATE{$s \leftarrow greedyMutation (bestSequence,targetStructure)$}
\ELSE
\STATE{$s \leftarrow randomMutation (bestSequence,targetStructure)$}
\ENDIF
\STATE{Update $bestSequence$ with $s$ using Multi Objective comparison}
\ENDWHILE
\end{ALC@g}
\end{algorithmic}
\end{algorithm}

\subsection{Progressive Narrowing}

PN is an improvement on MOGRLS that starts searching multiple sequences before focusing the search on the best one. It is described in Algorithm \ref{AlgoPN}.

\begin{algorithm}
\caption{Progressive Narrowing}
\begin{algorithmic}[1]
\label{AlgoPN}
\STATE{PN ($targetStructure$)}
\begin{ALC@g}
\FOR{number of restarts}
\STATE{$s \leftarrow greedyMutation (bestSequences [n],targetStructure)$}
\STATE{$nevals \leftarrow [0,...,0]$}
\WHILE{True}
\FOR{$n \in range(len(bestSequences))$}
\IF{$nevals [n] < 500$}
\STATE{$s \leftarrow greedyMutation (bestSequences [n],targetStructure)$}
\ELSE
\STATE{$s \leftarrow randomMutation (bestSequences [n],targetStructure)$}
\ENDIF
\STATE{Update $bestSequences [n]$ with $s$ using Multi Objective comparison}
\STATE{$nevals [n] \leftarrow nevals [n] + 1$}
\ENDFOR
\FOR{number of best sequences}
\IF{$profile [n] == nevals [n]$}
\STATE{remove worst sequence from best sequences}
\STATE{remove $profile [n]$ from profile}
\ENDIF
\ENDFOR
\STATE{break if one sequence left and $profile [0] == nevals [0]$}
\ENDWHILE
\ENDFOR
\end{ALC@g}
\end{algorithmic}
\end{algorithm}

\subsection{Search for Parameter Tuning}

The search for the parameters of PN is done using Algorithm \ref{AlgoParameterGeneration} for generating the possible combinations of parameters that sum to a predefined number of evaluations, and Algorithm \ref{AlgoParameterTuning} for testing on a dataset of previous recorded runs of MOGRLS the score of each combination of parameters. 

\begin{algorithm}
\caption{Parameter generation.}
\begin{algorithmic}[1]
\label{AlgoParameterGeneration}
\STATE{search ($n, k, s, current, l, possible$)}
\begin{ALC@g}
\IF{$k == 0$}
\IF{$s == n$}
\STATE{$l.append (current)$}
\RETURN{}
\ENDIF
\ENDIF
\IF{$s \geq n$}
\RETURN{}
\ENDIF
\STATE{$start \leftarrow 0$}
\IF{$len(current) > 0$}
\STATE{$start \leftarrow current [-1]$}
\ENDIF
\FOR{$i \in possible$}
\IF{$i \geq start$}
\STATE{$si \leftarrow s + k \times (i - start)$}
\STATE{$cur \leftarrow copy(current)$}
\STATE{$cur.append (i)$}
\STATE{search ($n, k - 1, si, cur, l, possible$)}
\ENDIF
\ENDFOR
\end{ALC@g}
\end{algorithmic}
\end{algorithm}

\begin{algorithm}
\caption{Parameter tuning.}
\begin{algorithmic}[1]
\label{AlgoParameterTuning}
\STATE{ParameterTuning ()}
\begin{ALC@g}
\FOR{$restart \in [1350,2700]$}
\FOR{$n \in [1..maxprofile + 1]$}
\STATE{$targetSum \leftarrow restart$}
\STATE{$result \leftarrow []$}
\STATE{$search (targetSum, n, 0, [], result, possible)$}
\FOR{$strategy \in result$}
\STATE{$nbSolved \leftarrow 0$}
\FOR{$j \in [0..nsamples]$}
\STATE{$solved \leftarrow False$}
\FOR{$r \in [0..2700//restart]$}
\STATE{$quad \leftarrow sample(range(nprocess), len (strategy))$}
\FOR{$i \in [0..len (strategy) - 1]$}
\STATE{$index \leftarrow max (0, strategy [i] - 1)$}
\STATE{remove element with worst score after index evaluations in $quad$}
\ENDFOR
\ENDFOR
\IF{solved}
\STATE{$nbSolved \leftarrow nbSolved + 1$}
\ENDIF
\ENDFOR
\IF{$nbSolved > best$}
\STATE{$best \leftarrow nbSolved$}
\STATE{memorize $strategy$}
\ENDIF
\ENDFOR
\ENDFOR
\ENDFOR
\end{ALC@g}
\end{algorithmic}
\end{algorithm}

\subsection{Multi Objective Generalized Nested Rollout Policy Adaptation with Limited Repetitions}

This section presents the MOGNRPALR algorithms which is a combinations of GNRPA \cite{Cazenave2020GNRPA}, GNRPALR \cite{cazenave2024generalized} and the Multi Objective evaluations of GREED-RNA \cite{lozano2024simple}. GNRPA is a generalization of the NRPA algorithm to the use of a prior. GNRPALR is an improvement of GNRPA that avoids too deterministic policies. It stops the iterations at a level when the best sequence of this level is found a second time. All of these algorithms are improvements of the Nested Rollout Policy Adaptation (NRPA) \cite{Rosin2011} algorithm whihc is an effective combination of NMCS and the online learning of a playout policy. 

In MOGNRPALR each move is associated to a weight stored in an array called the policy. For each level there is a best sequence and a policy. The principle is to reinforce the weights of the best sequence of moves found during the iterations at each level. At the lowest level, the weights are used in the softmax function to produce a playout policy that generates good sequences of moves. 

MOGNRPALR use nested search \cite{CazenaveIJCAI09}. At each level it takes a policy as input and returns a sequence and its associated scores. At any level $>$ 0, the algorithm makes numerous recursive calls to the lower level, adapting the policy each time with the best sequence of moves to date. The changes made to the policy do not affect the policy in higher levels. At level 0, MOGNRPALR return the sequence obtained by the playout function as well as its associated scores.

The playout function sequentially constructs a random solution biased by the weights of the moves until it reaches a terminal state. At each step, the function performs Boltzmann sampling, choosing the actions with a probability given by the softmax function.

Let $w_{m}$ be the weight associated to a move $m$ in the policy. In NRPA, the probability of choosing move $m$ is defined by: 

$$ p_{m} = \frac{e^{w_{m}}}{\sum_k{e^{w_{k}}}} $$

where $k$ goes through the set of possible moves, including $m$.

GNRPA \cite{Cazenave2020GNRPA} generalizes the way probability is calculated using bias $\beta_{m}$. The probability of choosing the move $m$ becomes: 
$$ p_{m} = \frac{e^{w_{m}+\beta_{m}}}{\sum_k{e^{w_{k}+\beta_{k}}}} $$

Taking $\beta_{m} = \beta_{k} = 0$, we find the formula for NRPA again which corresponds to sampling without prior. 

The algorithm for performing playouts in MOGNRPALR is given in algorithm \ref{PLAYOUT}. The main MOGNRPALR algorithm is given in the algorithm \ref{MOGNRPALR}. MOGNRPALR calls the adapt algorithm to modify the policy weights so as to reinforce the best sequence of the current level. The policy is passed by reference to the adapt algorithm which is given in the algorithm \ref{ADAPT}.

The principle of the adapt function is to increase the weights of the moves of the best sequence of the level and to decrease the weights of all possible moves by an amount proportional to their probabilities of being played. $\delta_{bm} = 0$ when $b \neq m$ and $\delta_{bm} = 1$ when $b = m$.

At line 16 of Algorithm \ref{MOGNRPALR} there is a condition that corresponds to Stabilized NRPA \cite{Cazenave2020Stabilized} and to starting to adapt only after a few iterations \cite{cazenave2016selective,Cazenave2020Inverse,dang2023warm}.

\begin{algorithm}
\begin{algorithmic}[1]
\STATE{playout ($policy$)}
\begin{ALC@g}
\STATE{$state \leftarrow root$}
\STATE{$sequence \leftarrow []$}
\WHILE{true}
\IF{terminal($state$)}
\RETURN{scores ($state$), $sequence$}
\ENDIF
\STATE{$z$ $\leftarrow$ 0}
\FOR{$m \in$ possible moves for $state$}
\STATE{$o [m] \leftarrow e^{policy[code(m)] + \beta_m}$}
\STATE{$z \leftarrow z + o [m]$}
\ENDFOR
\STATE{choose a $move$ with probability $\frac{o [move]}{z}$}
\STATE{play ($state$, $move$)}
\STATE{$sequence$.append ($move$)}
\ENDWHILE
\end{ALC@g}
\end{algorithmic}
\caption{\label{PLAYOUT}The Multi Objective playout algorithm}
\end{algorithm}

\begin{algorithm}
\begin{algorithmic}[1]
\STATE{adapt ($policy$, $sequence$)}
\begin{ALC@g}
\STATE{$polp \leftarrow policy$}
\STATE{$state \leftarrow root$}
\FOR{$b \in sequence$}
\STATE{$z \leftarrow 0$}
\FOR{$m \in$ possible moves for $state$}
\STATE{$o [m] \leftarrow e^{policy[code(m)] + \beta_m}$}
\STATE{$z \leftarrow z + o [m]$}
\ENDFOR
\FOR{$m \in$ possible moves for $state$}
\STATE{$p_m \leftarrow \frac{o[m]}{z}$}
\STATE{$polp [code(m)] \leftarrow polp [code(m)] - \alpha(p_m - \delta_{bm})$}
\ENDFOR
\STATE{play ($state$, $b$)}
\ENDFOR
\RETURN{$polp$}
\end{ALC@g}
\end{algorithmic}
\caption{\label{ADAPT}The adapt algorithm}
\end{algorithm}

\begin{algorithm}
\begin{algorithmic}[1]
\STATE{MOGNRPALR ($level$, $policy$)}
\begin{ALC@g}
\IF{level == 0}
\RETURN{playout ($policy$)}
\ELSE
\STATE{$bestSequences \leftarrow []$}
\STATE{$i \leftarrow 0$}
\WHILE{True}
\STATE{$scores$,$new$ $\leftarrow$ MOGNRPALR($level-1$, $policy$)}
\IF{$bestSequences \neq$ []}
\IF{$new$ == $bestSequences$[0][1]}
\RETURN{$scores$, $new$}
\ENDIF
\ENDIF
\STATE{$bestSequences$.append ([$scores$,$new$])}
\STATE{$bestSequences$.sort ()}
\IF{$level > 2~\OR~level < 3~and~i > 3~\OR~level == 1~and~i > 3~and~i \% 4 == 0$}
\STATE{$policy \leftarrow$ adapt ($policy$, $bestSequences$[0][1])}
\ENDIF
\STATE{$i \leftarrow i + 1$}
\ENDWHILE
\ENDIF
\end{ALC@g}
\end{algorithmic}
\caption{\label{MOGNRPALR}The MOGNRPALR algorithm.}
\end{algorithm}

\section{Experimental Results}

\begin{figure}[h]
    \centering
    \begin{subfigure}{0.32\textwidth}
        \centering
        \includegraphics[width=1cm]{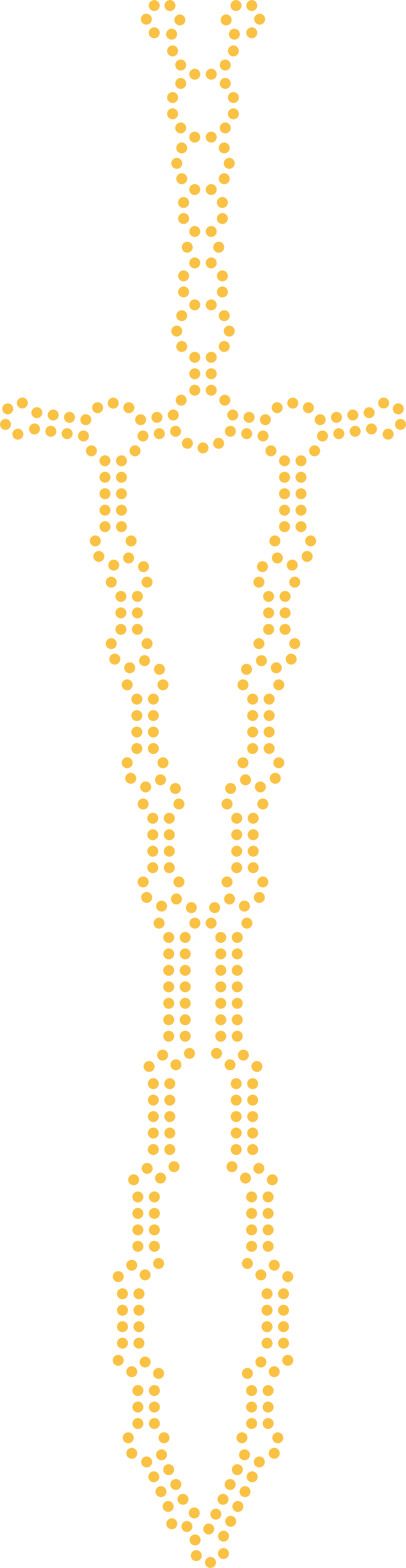}
        \caption{Gladius: problem 90}
        \label{fig:sub1}
    \end{subfigure}
    \hfill
    \begin{subfigure}{0.32\textwidth}
        \centering
        \includegraphics[width=3cm]{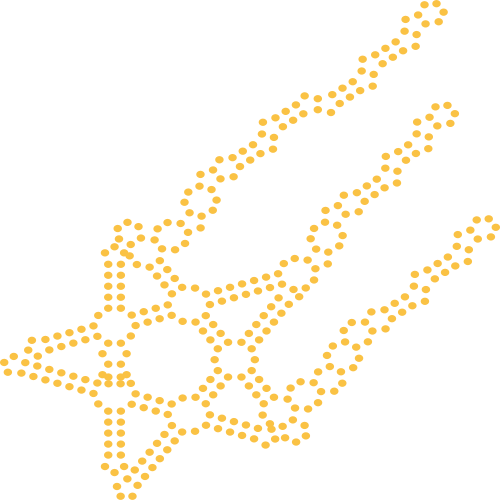}
        \caption{Shooting Star: problem 99}
        \label{fig:sub2}
    \end{subfigure}
    \begin{subfigure}{0.32\textwidth}
        \centering
        \includegraphics[width=3cm]{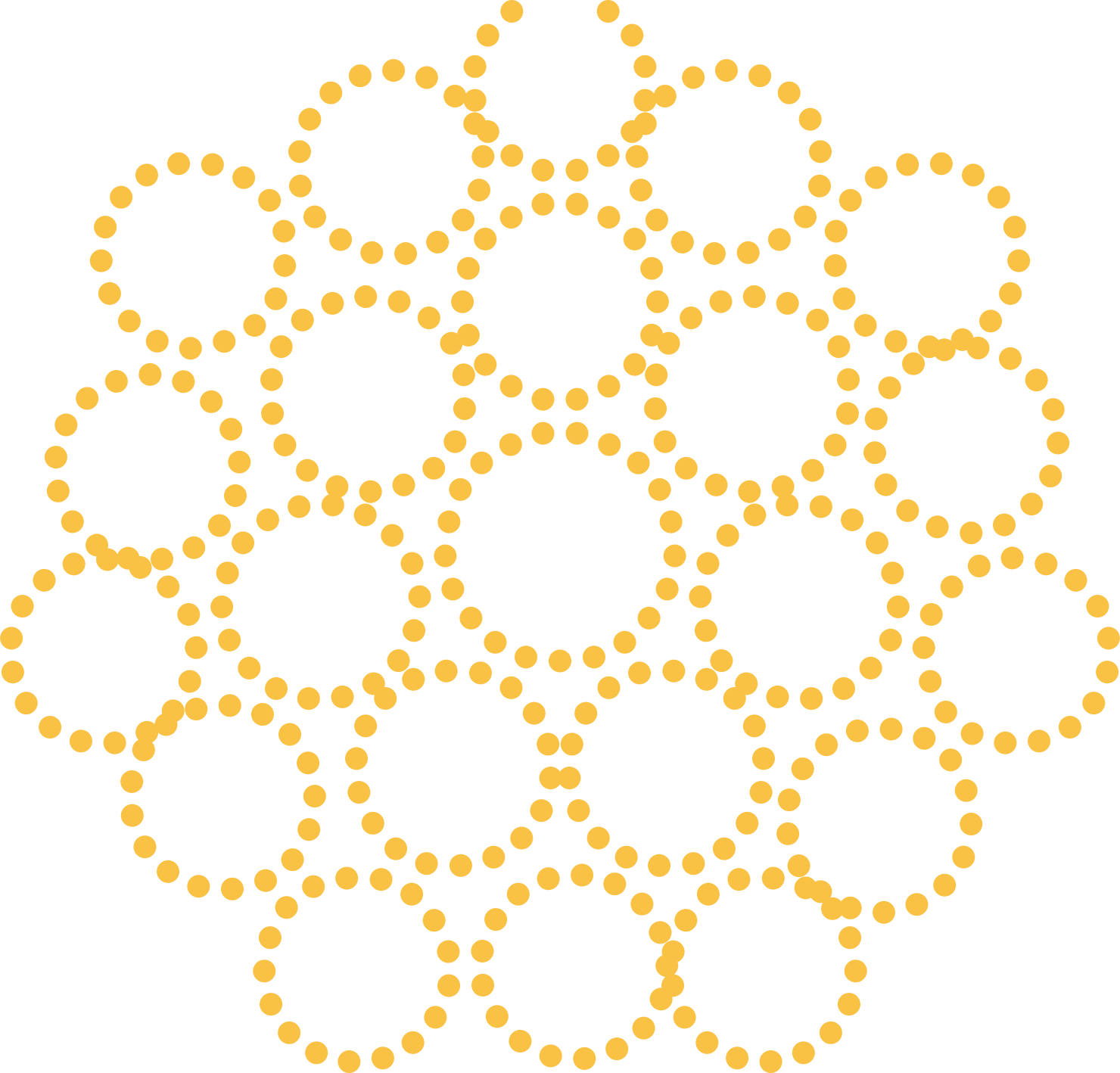}
        \caption{Teslagon: problem 100}
        \label{fig:sub2}
    \end{subfigure}
    \caption{Problems 90, 99 and 100 from Eterna100 v1.}
    \label{eterna}
\end{figure}

The same default parameters are always used for both GREED-RNA and MOGNRPALR. The biases used in MOGNRPALR are 5.0 for GC, CG and A and 0.0 for AU, UA, UG, GU, C, G and U. The Turner 1999 parameters are used for RNA folding. The same Multi Objective evaluations are used for all algorithms. The problems we solve are the original Eterna100 v1 problems. Running 200 processes in parallel MOGNRPALR solves all the problems of Eterna100 v1 in less than one day. In the following, we will focus on three of the most difficult problems: Problems 90, 99 and 100. Figure \ref{eterna} gives the target secondary structures for problems 90, 99 and 100.

\subsection{MOGRLS}

\begin{figure}[h]
    \centering
    \begin{subfigure}{0.45\textwidth}
        \centering
        \includegraphics[width=\linewidth]{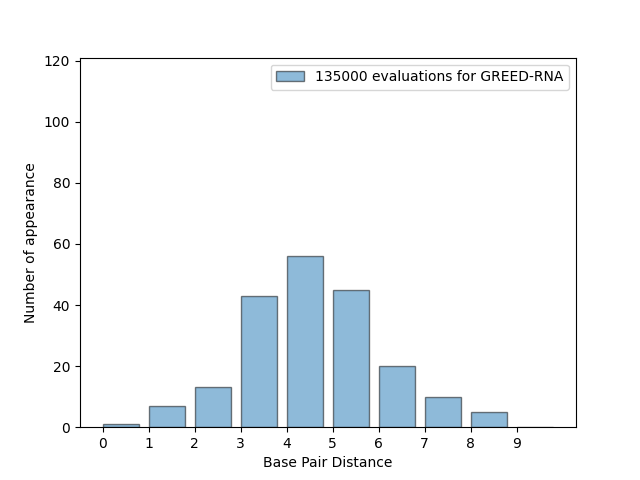}
        \caption{Distribution of the BPD for problem 99 after 135 000 evaluations by GREED-RNA.}
        \label{fig:sub1}
    \end{subfigure}
    \hfill
    \begin{subfigure}{0.45\textwidth}
        \centering
        \includegraphics[width=\linewidth]{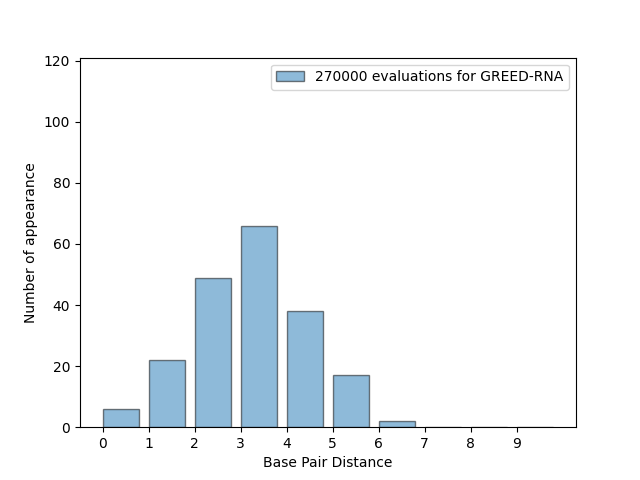}
        \caption{Distribution of the BPD for problem 99 after 270 000 evaluations by GREED-RNA.}
        \label{fig:sub1}
    \end{subfigure}
    \hfill
    \begin{subfigure}{0.45\textwidth}
        \centering
        \includegraphics[width=\linewidth]{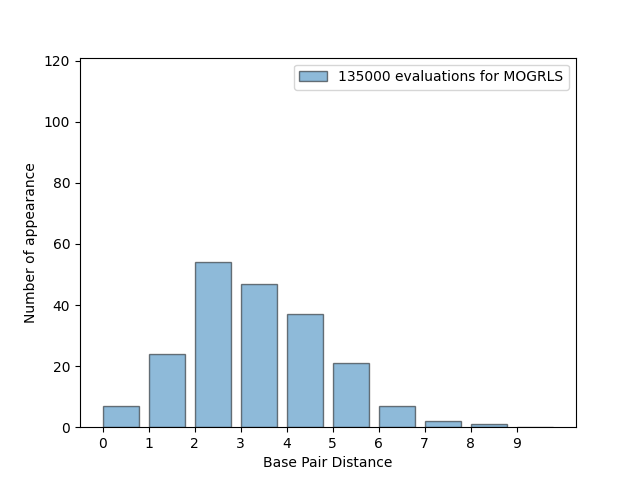}
        \caption{Distribution of the BPD for problem 99 after 135 000 evaluations by MOGRLS.}
        \label{fig:sub2}
    \end{subfigure}
    \hfill
    \begin{subfigure}{0.45\textwidth}
        \centering
        \includegraphics[width=\linewidth]{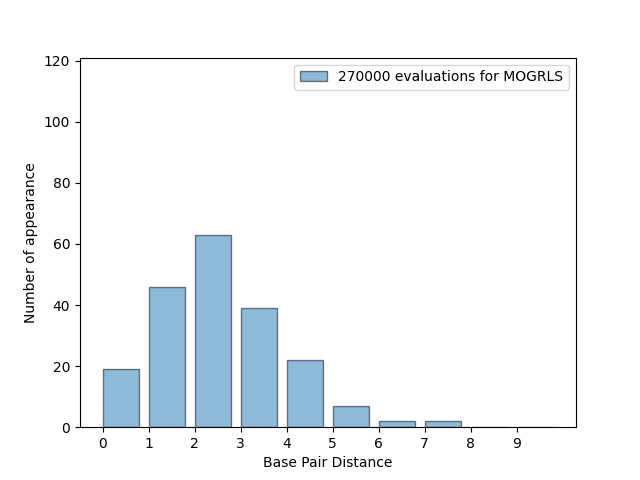}
        \caption{Distribution of the BPD for problem 99 after 270 000 evaluations by MOGRLS.}
        \label{fig:sub1}
    \end{subfigure}
    \hfill
    \begin{subfigure}{0.45\textwidth}
        \centering
        \includegraphics[width=\linewidth]{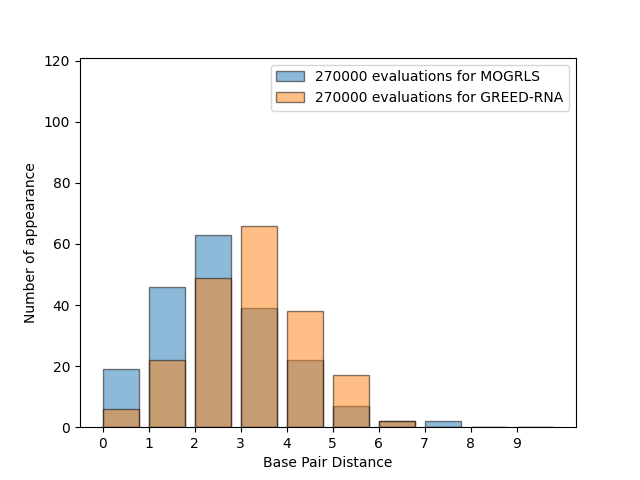}
        \caption{Comparison after 270 000 evaluations between GREED-RNA and MOGRLS.}
        \label{fig:sub1}
    \end{subfigure}
    \hfill
    \begin{subfigure}{0.45\textwidth}
        \centering
        \includegraphics[width=\linewidth]{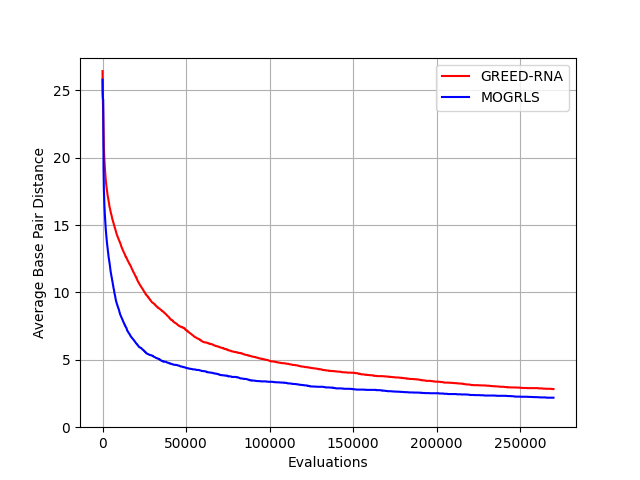}
        \caption{Evolution of the average BPD of GREED-RNA and MOGRLS.}
        \label{fig:sub2}
    \end{subfigure}
    \caption{Comparison of the distributions of BPD on problem 99 for GREED-RNA and MOGRLS for increasing numbers of evaluations. After 270 000 evaluations which corresponds to one day of computation, MOGRLS has solved the problem 19 times out of 200 runs while GREED-RNA has solved the problem 2 times out of 200 runs (see subfigure (c)). Subfigure (d) gives the evolution of the average BPD with the number of evaluations for problem 99. The averages are calculated using 200 runs. The rightmost value corresponds to one day of computation for one run.}
    \label{MOGRLS}
\end{figure}

Figure \ref{MOGRLS} gives the evolution of the BPD for problem 99 both for GREED-RNA and MOGRLS. We can see that MOGRLS gets better results.

\subsection{PN}

\begin{figure}[h]
    \centering
    \begin{subfigure}{0.45\textwidth}
        \centering
        \includegraphics[width=\linewidth]{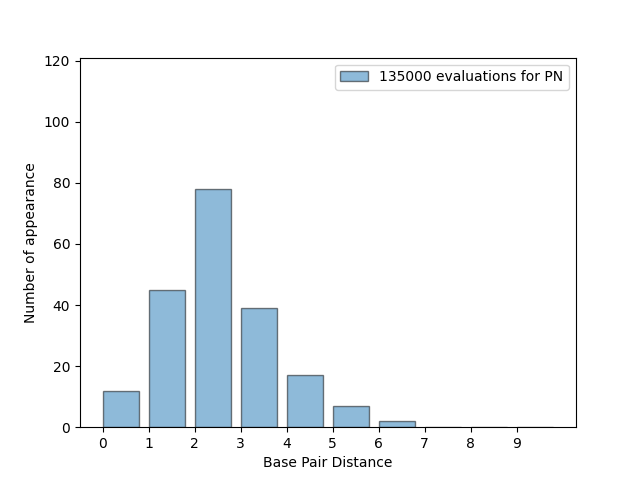}
        \caption{Distribution of the BPD for problem 99 after 135 000 evaluations by PN.}
        \label{fig:sub1}
    \end{subfigure}
    \hfill
    \begin{subfigure}{0.45\textwidth}
        \centering
        \includegraphics[width=\linewidth]{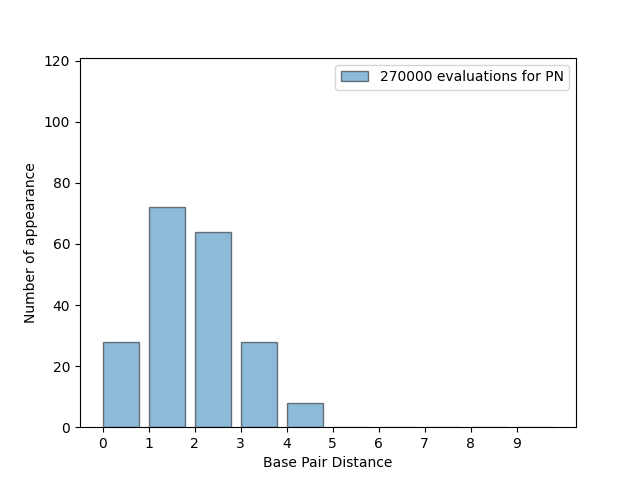}
        \caption{Distribution of the BPD for problem 99 after 270 000 evaluations by PN.}
        \label{fig:sub2}
    \end{subfigure}
    \hfill
    \begin{subfigure}{0.45\textwidth}
        \centering
        \includegraphics[width=\linewidth]{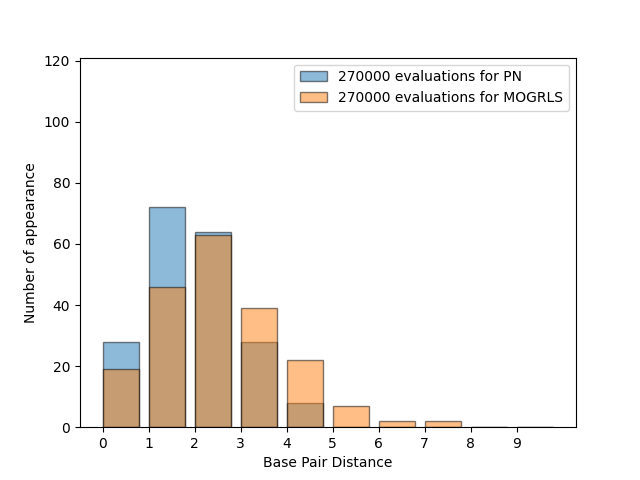}
        \caption{Comparison after 270 000 evaluations between PN and MOGRLS.}
        \label{fig:sub1}
    \end{subfigure}
    \hfill
    \begin{subfigure}{0.45\textwidth}
        \centering
        \includegraphics[width=\linewidth]{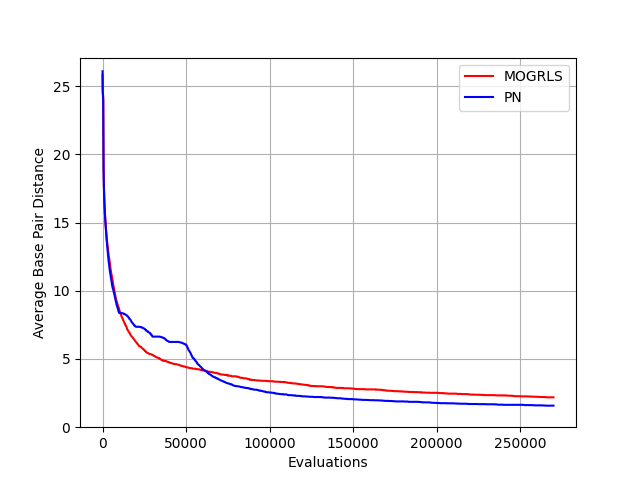}
        \caption{Evolution of the average BPD of PN and MOGRLS for problem 99.}
        \label{fig:sub2}
    \end{subfigure}
    \caption{Comparison of the distributions of BPD on problem 99 for PN and MOGRLS for increasing numbers of evaluations. After 270 000 evaluations which corresponds to one day of computation, PN has solved the problem 28 times out of 200 runs while MOGRLS has solved the problem 19 times out of 200 runs (see subfigure (c)). Subfigure (d) gives the evolution of the average BPD with the number of evaluations for problem 99. The averages are calculated using 200 runs. The rightmost value corresponds to one day of computation for one run.}
    \label{PN}
\end{figure}

The results for MOGRLS of the previous section have been stored in order to be used as a dataset. For all of the 200 processes, the best BPD of each process has been stored every 100 evaluations. The search for narrowing profiles was done for restarts of 135 000 and 270 000 evaluations. The restart of 270 000 evaluations corresponds to no restart. For each profile tested, 100 000 combinations were tested and averaged. The best profile found is [10 000, 10 000, 10 000, 10 000, 230 000] with no restart. It solves 16.77 \% of the combinations. Figure \ref{PN} gives the results for PN on problem 99 and compares them with MOGRLS. PN gives slightly better results than MOGRLS.

\subsection{MOGNRPALR}

Figure \ref{MOGNRPALR} gives the evolution of the BPD for problem 99 both for MOGNRPALR. It also compares MOGNRPALR with GREED-RNA, PN, and MOGRLS. We can see that MOGNRPALR gets much better results.

\begin{figure}[h]
    \centering
    \begin{subfigure}{0.45\textwidth}
        \centering
        \includegraphics[width=\linewidth]{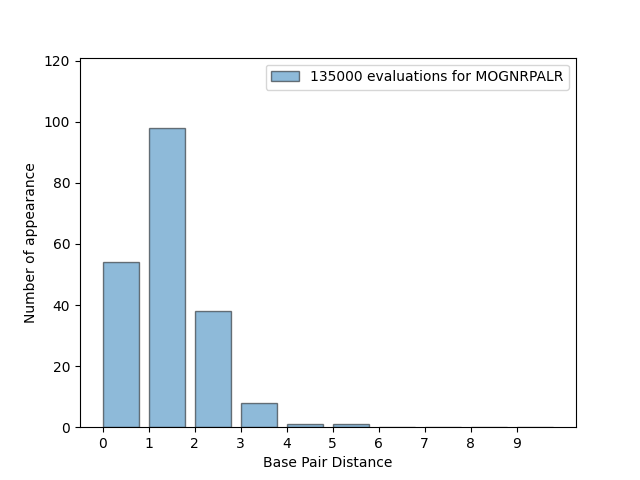}
        \caption{Distribution of the BPD for problem 99 after 135 000 evaluations by MOGNRPALR.}
        \label{fig:sub1}
    \end{subfigure}
    \hfill
    \begin{subfigure}{0.45\textwidth}
        \centering
        \includegraphics[width=\linewidth]{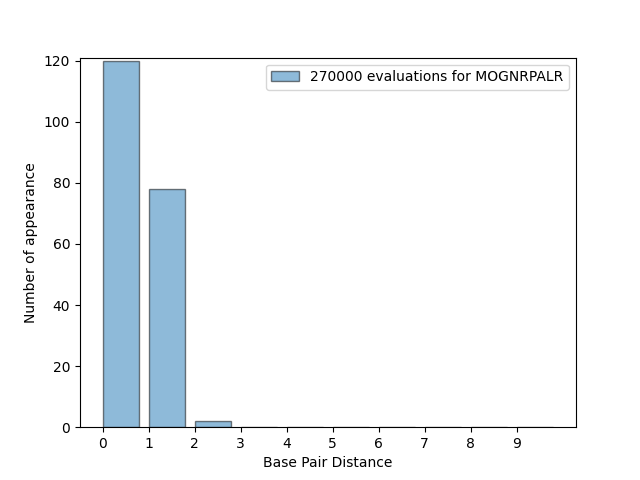}
        \caption{Distribution of the BPD for problem 99 after 270 000 evaluations by MOGNRPALR.}
        \label{fig:sub2}
    \end{subfigure}
    \hfill
    \begin{subfigure}{0.45\textwidth}
        \centering
        \includegraphics[width=\linewidth]{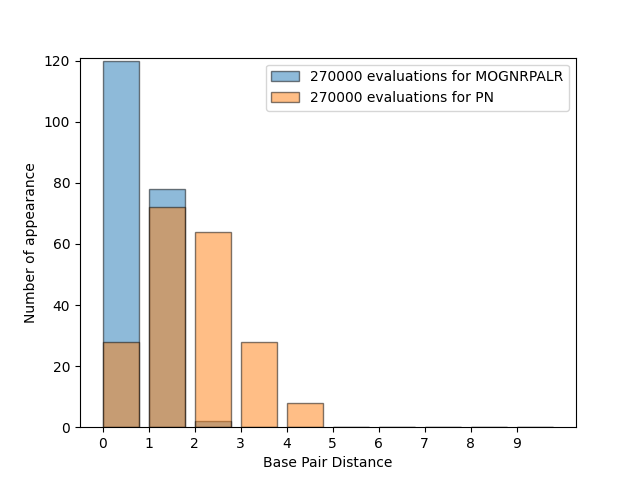}
        \caption{Comparison after 270 000 evaluations between MOGNRPALR and PN.}
        \label{fig:sub1}
    \end{subfigure}
    \hfill
    \begin{subfigure}{0.45\textwidth}
        \centering
        \includegraphics[width=\linewidth]{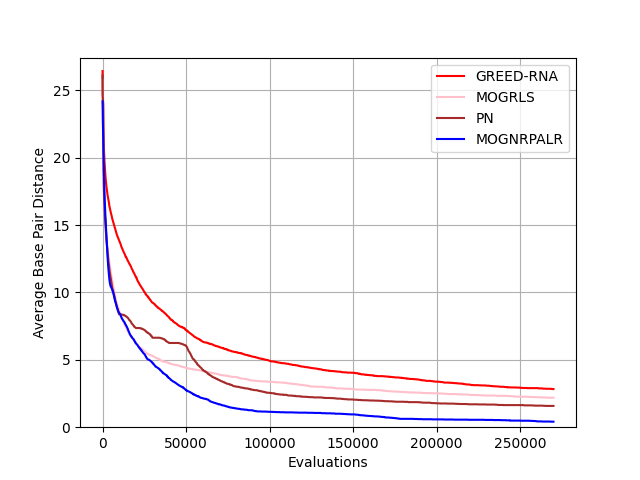}
        \caption{Evolution of the average BPD of all algorithms for problem 99.}
        \label{fig:sub2}
    \end{subfigure}
    \caption{Comparison of the distributions of BPD on problem 99 for PN and MOGNRPALR for increasing numbers of evaluations. After 270 000 evaluations which corresponds to one day of computation, MOGNRPALR has solved the problem 120 times out of 200 runs while PN has solved the problem 28 times out of 200 runs (see subfigure (c)). Subfigure (d) gives the evolution of the average BPD with the number of evaluations for problem 99 for all algorithms. The averages are calculated using 200 runs. The rightmost value corresponds to one day of computation for one run.}
    \label{MOGNRPALR}
\end{figure}

Table \ref{table} gives the distributions of BPD after 270 000 evaluations for problem 99. The algorithms are sorted by increasing performances. GREED-RNA solves the problem 6 times out of 200 while MOGNRPALR solves it 120 times out of 200.

\begin{table}[h]
\caption{Distributions of the BPD of the various algorithms after 270 000 evaluations for problem 99.\label{table}}
\centering
\begin{tabular}{lrrrrrrrrr}
\hline
BPD&0&1&2&3&4&5&6&7&8\\
\hline
\texttt{GREED-RNA} & 6 & 22 & 49 & 66 & 38 & 17 & 2 &  0 & 0\\
\texttt{MOGRLS} & 19 & 46 & 63 & 39 & 22 & 7 & 2 &  2 & 0\\
\texttt{PN} & 28 & 72 & 64 & 28 & 8 & 0 & 0 & 0 & 0\\
\texttt{MOGNRPALR} & ~120 & ~~78 & ~~~2 & ~~~0 & ~~~0 & ~~~0 & ~~~0 & ~~~0 & ~~~0\\
\hline
\end{tabular}
\end{table}

\subsection{Problem 90}

Problem 90 is a difficult problem where the behavior of the search algorithms is a slow descent toward the lower BPD.

The number of evaluations performed by GREED-RNA in one hour for problem 90 is 9 200. So we compare GREED-RNA to MOGNRPALR after 220 000 evaluations, which corresponds to one day of computation. MOGNRPALR reaches 10 400 in one hour for problem 90. The comparison using the same number of evaluations for the two algorithms is slightly favorable for GREED-RNA.

\begin{figure}[h]
    \centering
    \begin{subfigure}{0.45\textwidth}
        \centering
        \includegraphics[width=\linewidth]{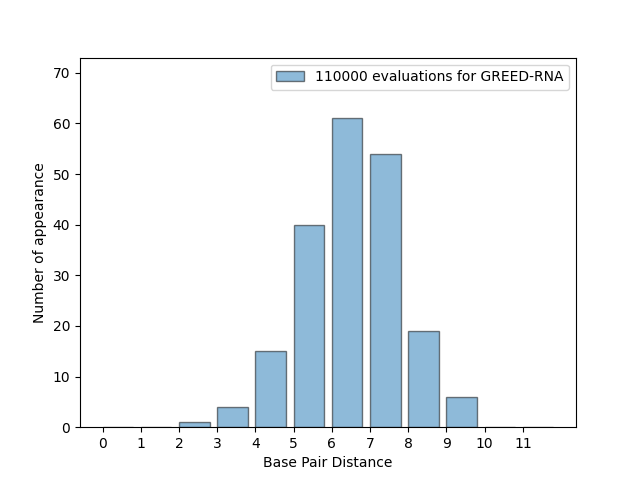}
        \caption{Distribution of the BPD for problem 90 after 110 000 evaluations by GREED-RNA.}
        \label{fig:sub1}
    \end{subfigure}
    \hfill
    \begin{subfigure}{0.45\textwidth}
        \centering
        \includegraphics[width=\linewidth]{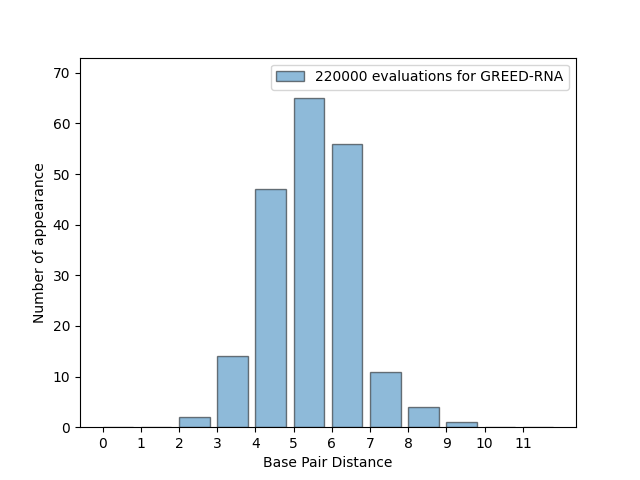}
        \caption{Distribution of the BPD for problem 90 after 220 000 evaluations by GREED-RNA.}
        \label{fig:sub2}
    \end{subfigure}
    \centering
    \begin{subfigure}{0.45\textwidth}
        \centering
        \includegraphics[width=\linewidth]{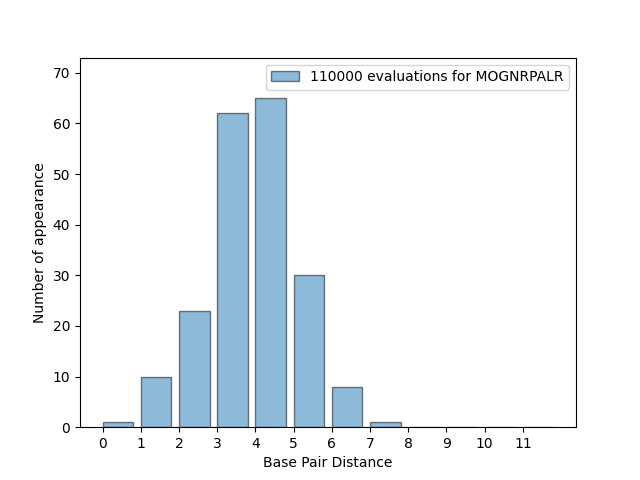}
        \caption{Distribution of the BPD for problem 90 after 110 000 evaluations by MOGNRPALR.}
        \label{fig:sub1}
    \end{subfigure}
    \hfill
    \begin{subfigure}{0.45\textwidth}
        \centering
        \includegraphics[width=\linewidth]{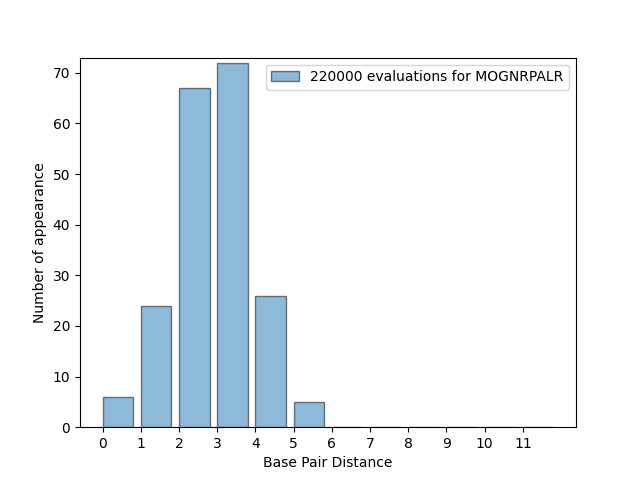}
        \caption{Distribution of the BPD for problem 90 after 220 000 evaluations by MOGNRPALR.}
        \label{fig:sub2}
    \end{subfigure}
    \hfill
    \begin{subfigure}{0.45\textwidth}
        \centering
        \includegraphics[width=\linewidth]{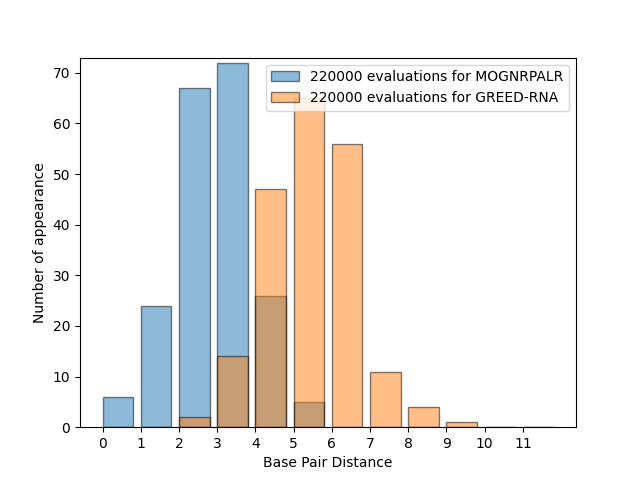}
        \caption{Comparison of the distributions after 220 000 evaluations between GREED-RNA and MOGNRPALR.}
        \label{fig:sub1}
    \end{subfigure}
    \hfill
    \begin{subfigure}{0.45\textwidth}
        \centering
        \includegraphics[width=\linewidth]{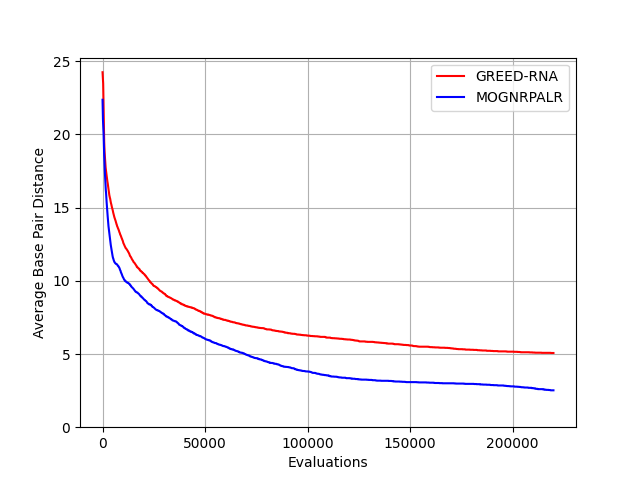}
        \caption{Evolution of the average BPD of GREED-RNA and MOGNRPALR for problem 90.}
        \label{fig:sub2}
    \end{subfigure}
    \caption{Comparison of the BPD on problem 90 for GREED-RNA and MOGNRPALR for increasing numbers of evaluations. 220 000 evaluations by one process takes one day. GREED-RNA is stuck and does not solve the problem while MOGNRPALR progresses and solves the problem 6 times out of 200 runs.}
    \label{pb90}
\end{figure}

Figure \ref{pb90} gives the distributions of BPD for GREED-RNA and MOGNRPALR. None of the 200 GREED-RNA processes finds a solution, and the best BPD found is 2. MOGNRPALR finds multiple solutions. 

\subsection{Problem 100}

Problem 100 is also a difficult problem. The number of evaluations performed by GREED-RNA in one hour for problem 100 is 37 000. So we compare GREED-RNA to MOGNRPALR after 530 000 evaluations, which corresponds to one day of computation.

\begin{figure}[h]
    \centering
    \begin{subfigure}{0.45\textwidth}
        \centering
        \includegraphics[width=\linewidth]{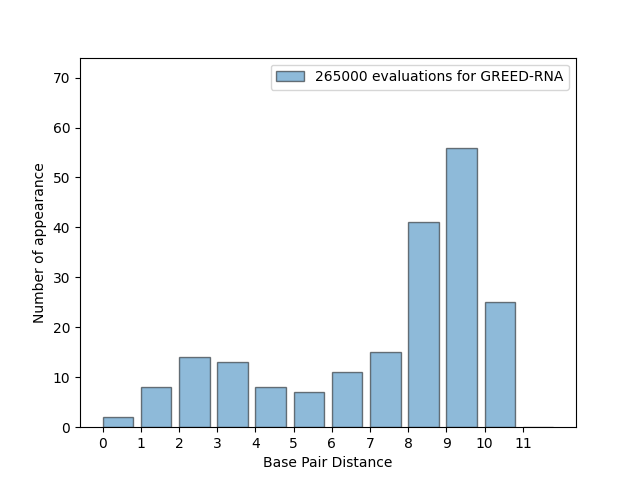}
        \caption{Distribution of the BPD for problem 100 after 265 000 evaluations by GREED-RNA.}
        \label{fig:sub1}
    \end{subfigure}
    \hfill
    \begin{subfigure}{0.45\textwidth}
        \centering
        \includegraphics[width=\linewidth]{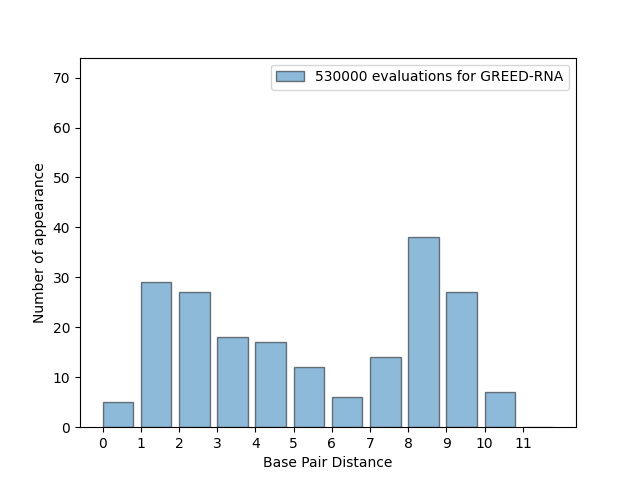}
        \caption{Distribution of the BPD for problem 100 after 530 000 evaluations by GREED-RNA.}
        \label{fig:sub2}
    \end{subfigure}
    \centering
    \begin{subfigure}{0.45\textwidth}
        \centering
        \includegraphics[width=\linewidth]{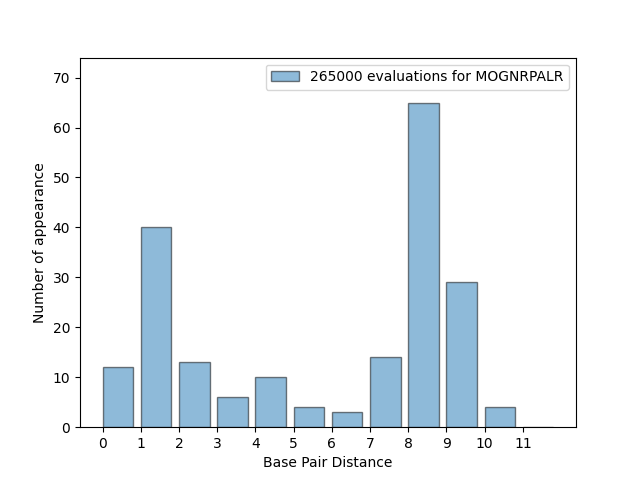}
        \caption{Distribution of the BPD for problem 100 after 265 000 evaluations by MOGNRPALR.}
        \label{fig:sub1}
    \end{subfigure}
    \hfill
    \begin{subfigure}{0.45\textwidth}
        \centering
        \includegraphics[width=\linewidth]{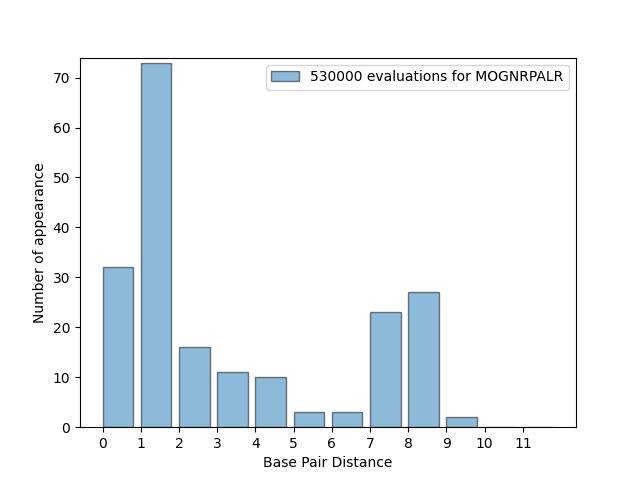}
        \caption{Distribution of the BPD for problem 100 after 530 000 evaluations by MOGNRPALR.}
        \label{fig:sub2}
    \end{subfigure}
    \hfill
    \begin{subfigure}{0.45\textwidth}
        \centering
        \includegraphics[width=\linewidth]{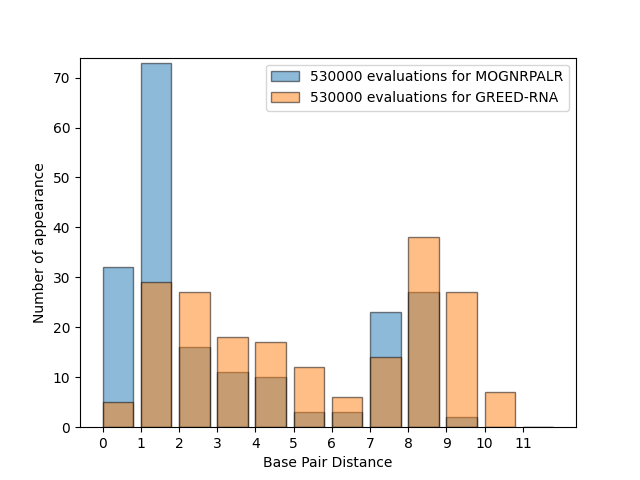}
        \caption{Comparison of the distributions after 530 000 evaluations between GREED-RNA and MOGNRPALR.}
        \label{fig:sub1}
    \end{subfigure}
    \hfill
    \begin{subfigure}{0.45\textwidth}
        \centering
        \includegraphics[width=\linewidth]{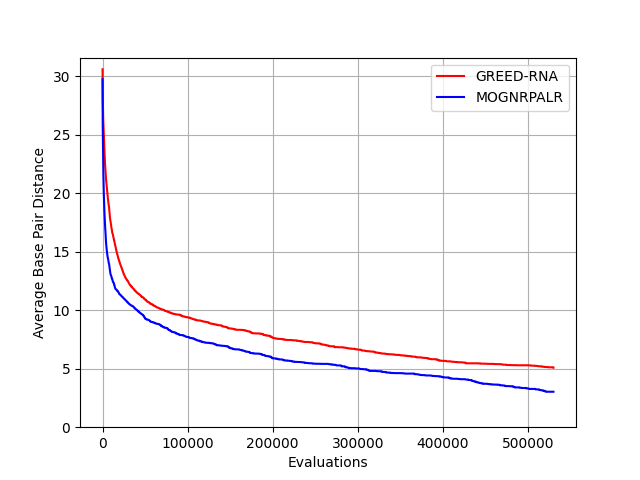}
        \caption{Evolution of the average BPD of GREED-RNA and MOGNRPALR for problem 100.}
        \label{fig:sub2}
    \end{subfigure}
    \caption{Comparison of the BPD on problem 100 for GREED-RNA and MOGNRPALR for increasing numbers of evaluations. 530 000 evaluations by one process takes one day. GREED-RNA solves problem 100 less frequently than MOGNRPALR.}
    \label{pb100}
\end{figure}

Figure \ref{pb100} gives the distributions of BPD for GREED-RNA and MOGNRPALR. MOGNRPALR finds many more solutions than GREED-RNA. 

\section{Conclusion}

Montparnasse is a framework for RNA design. It contains algorithms such as MOGRLS that simplify and improve the GREED-RNA local search approach to RNA design. It improves on MOGRLS with PN and automatic parameter tuning. The main result of this paper is the design and application of the MOGNRPALR algorithm to RNA design. For difficult problems of Eterna100 v1 it gives much better results than greedy local search. It solves the Eterna100 v1 benchmark since all problems are solved within one day using 200 runs in parallel, each run using the same number of evaluations as the number of evaluations of a single run in one day.